\documentclass{article}
\usepackage{spconf,amsmath,graphicx,hyperref}
\usepackage{marvosym}
\usepackage{cite}
\usepackage{amssymb}
\usepackage{cleveref}
\usepackage{enumitem}
\setlist[itemize]{leftmargin=*, nosep, noitemsep, topsep=0pt, partopsep=0pt}
\usepackage{booktabs}
\usepackage{cuted}
\usepackage{float}
\usepackage{array} 

\usepackage[numbers]{natbib}
\usepackage{caption}
\usepackage{multicol}
\usepackage{natbib}
\title{Trace Is In Sentences: Unbiased Lightweight \\ChatGPT-Generated Text Detector}
\name{Mo Mu$^{*}$, Dianqiao Lei$^{*}$, Chang Li$^{{\dag}}$
\thanks{\noindent This work is supported by Beijing Natural Science Foundation under Grant No. QY25048.}
\thanks{$^{*}$Equal contribution.}
\thanks{{\dag} Corresponding Author.}}
\address{$^{1}$ Tsinghua University, Beijing, China}
\begin{document}
%
\maketitle
\begin{abstract}
The widespread adoption of ChatGPT has raised concerns about its misuse, highlighting the need for robust detection of AI-generated text. Current word-level detectors are vulnerable to paraphrasing or simple prompts (PSP), suffer from biases induced by ChatGPT’s word-level patterns (CWP) and training data content, degrade on modified text, and often require large models or online LLM interaction. To tackle these issues, we introduce a novel task to detect both original and PSP-modified AI-generated texts, and propose a lightweight framework that classifies texts based on their internal structure, which remains invariant under word-level changes. Our approach encodes sentence embeddings from pre-trained language models and models their relationships via attention. We employ contrastive learning to mitigate embedding biases from autoregressive generation and incorporate a causal graph with counterfactual methods to isolate structural features from topic-related biases. Experiments on two curated datasets, including abstract comparisons and revised life FAQs, validate the effectiveness of our method.

\end{abstract}
\begin{keywords}
Text Classification, Deepfake Detection, Counterfactual Learning
\end{keywords}
\section{Introduction}
\label{sec:intro}

With the rise of ChatGPT \cite{schulman2022chatgpt}—an LLM that mimics human writing and performs comparably to experts across tasks—its misuse in generating responses, papers, and assignments has led to misinformation proliferation \cite{kreps2022all} and academic integrity issues \cite{else2023chatgpt}, urgently necessitating reliable AI-generated text detection.

\begin{figure}[t]
    \centering
    \includegraphics[width=1.0\linewidth]{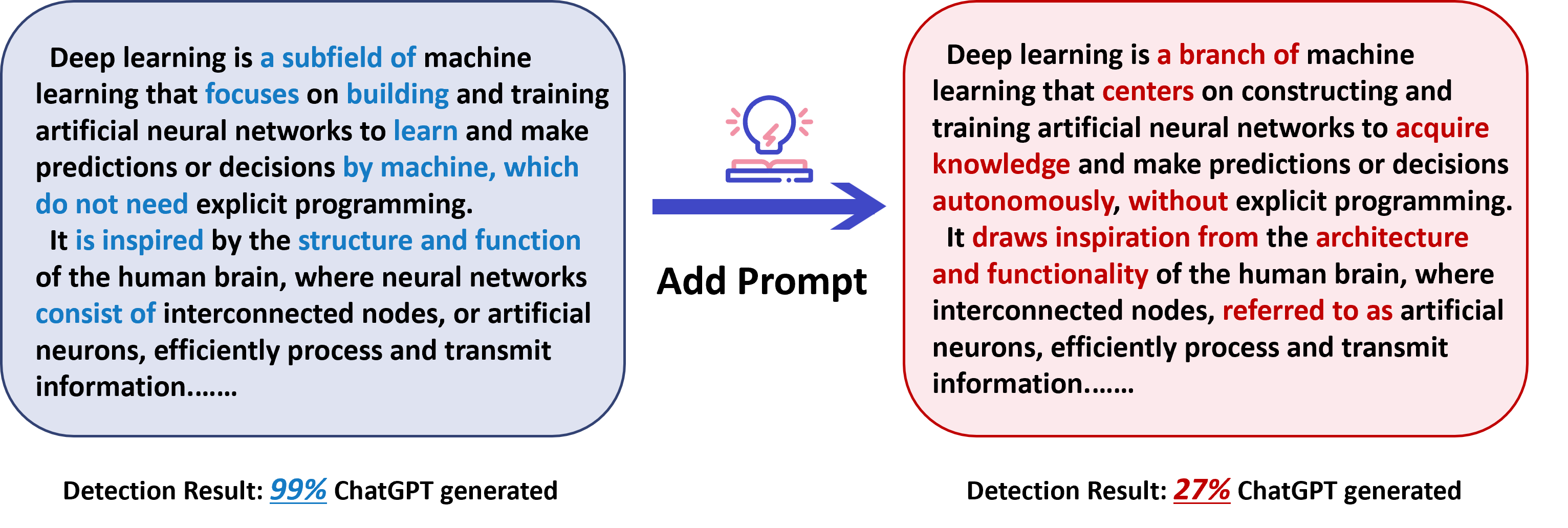}
    \vspace{-3mm}
    \caption{Motivation of our task: while existing detectors are often biased toward word-level patterns, making them vulnerable to simple adversarial attacks.}
    \label{fig:prompt}
    \vspace{-5mm}
\end{figure}

Existing detection approaches fall into two categories: one uses statistical metrics like perplexity, entropy, and rank \cite{gehrmann2019gltr} for threshold-based classification or regression \cite{verma2023ghostbuster}; however, closed-source LLMs limit the adaptability of models like GPT-4o \cite{hurst2024gpt}, and word-level sensitivity makes these methods fragile to minor edits \cite{liang2023gpt}. Another line employs Pretrained Langudge Models (PLMs) (e.g., RoBERTa \cite{ott2019fairseq}, BERT \cite{devlin2019bert}) for black-box classification based on word relations \cite{tian2023multiscale, guo2023close}. Yet, such models overlook structural cues and are vulnerable to paraphrasing or prompt-based polishing (PSP), leading to performance drops sharply under PSP attacks \cite{krishna2023paraphrasing} as shown in \cref{fig:prompt}.

We argue that text features include both intra- and inter-sentence levels, where PSP alters the former but hardly affects the latter\cite{lu2024less}. Prior methods may rely on spurious word-level patterns (CWP) from ChatGPT’s average stylistic bias, while humans exhibit individualized styles. Moreover, content bias in training data misleads classifiers. Hence, a robust detector should capture invariant structural relationships beyond surface words or topics.

\begin{figure*}[!t]
    \centering
    \includegraphics[width=0.75\linewidth]{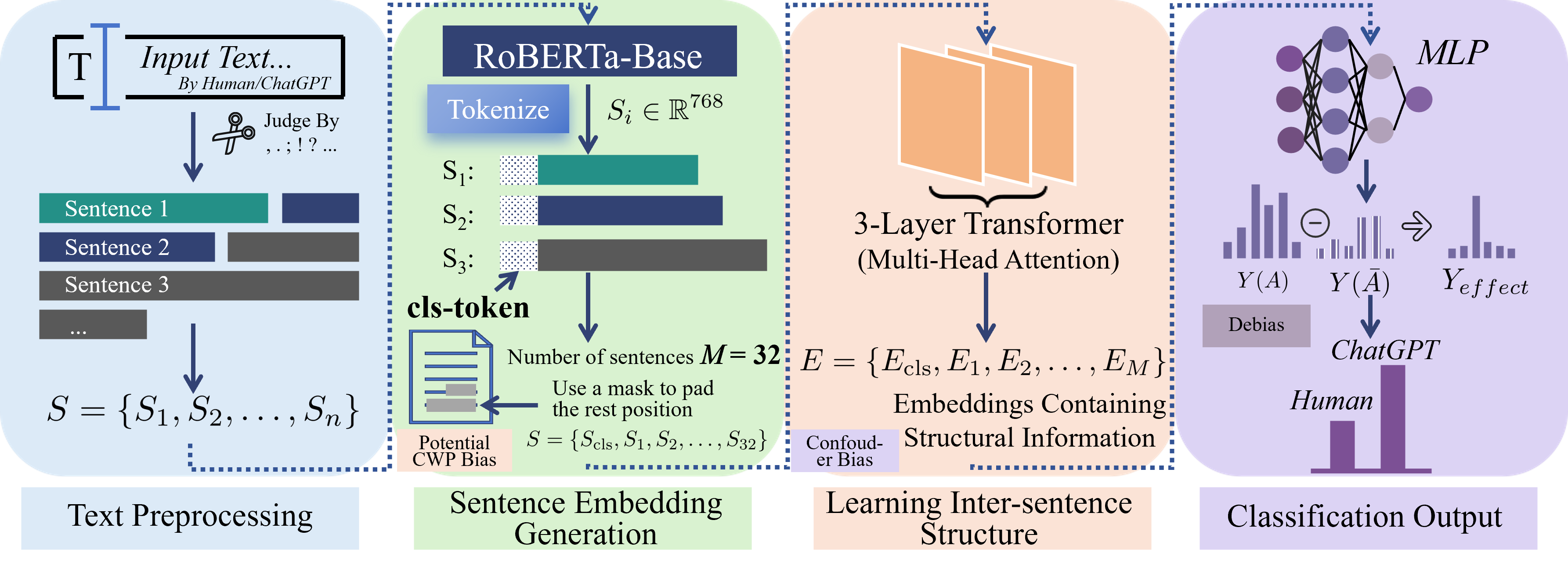}
    \caption{ Flowchart of Sentence-Relationship Extraction: Text split into sentences, embeddings via RoBERTa-base, fed to Transformer for inter-sentence structure learning, with CWP bias addressed counterfactually.}
    \label{fig:overview}
    \vspace{-5mm}
\end{figure*}

To address this, we propose a lightweight sentence-level relation detection framework. Using only sentence embeddings from PLMs and modeling their interactions via attention, we reduce sensitivity to word substitutions with lower parameter counts, ultimately improving adaptability. To counteract embedding bias from CWP, we apply contrastive learning with synonym-swapped machine text and machine-rewritten human text to enhance structural causality and suppress potential spurious correlations.

To evaluate PSP sensitivity and multi-domain performance, we contribute an abstract-comparison dataset (12,924 instances based on Arxiv and ChatGPT APIs \cite{gao2022comparing}) and multi-domain FAQ sets to evaluate robustness under PSP. Experiments show our model outperforms benchmarks in distinguishing AI-generated text by leveraging structural invariants.

Our contributions can be summarized as follows:
\begin{itemize}[nosep]
\item We identify and analyze the issue of word-level pattern bias in current ChatGPT-generated text detectors, and further examine it from a causal perspective.
\item Based on the causal graph we abstracted, we employ a lightweight detection head, which relies solely on robust inter-sentence structural relations and achieves effective detection performance.
\item We construct and release a large-scale benchmark including 263,595 English and 76,503 Chinese samples (based on HC3), enriched with cyclic translation, synonym substitution, and diverse prompts across multiple domains and at the same time validate the effectiveness and value of our method.

\end{itemize}

\section{Related Work}
\label{sec:format}

\subsection{AI-Generated Text Detection}
Since the release of GPT-2 \cite{radford2019language}, various strategies have been proposed to distinguish human-written from AI-generated text, which can be broadly categorized into zero-shot and fine-tuning-based methods.

Zero-shot methods typically leverage information from interactions with LLMs, such as directly querying whether a text is AI-generated, or utilizing features like perplexity (PPL) to capture differences in emotion, word choice, or sentence structure. GLTR \cite{gehrmann2019gltr} employs token-level statistics (e.g., perplexity, entropy, rank) from pre-trained models like BERT \cite{devlin2019bert} and GPT-2 for transparent and interpretable detection. DetectGPT \cite{mitchell2023detectgpt} assesses text authenticity by perturbing local wordings and evaluating global probability shifts. Ghostbuster \cite{verma2023ghostbuster} uses a feature set derived from multiple model outputs and mathematical operations to achieve strong out-of-domain performance.

Fine-tuning-based approaches often build on pre-trained language models. HuggingFace \cite{solaiman2019release} introduced an early GPT-2 detector based on RoBERTa \cite{ott2019fairseq}. Subsequent methods \cite{guo2023close, tian2023multiscale} leverage ChatGPT-generated datasets for improved detection. Closed-source detectors like GPTZero \cite{tian2023multiscale,abburi2023simple,habibzadeh2023gptzero} also show robust performance across diverse LLM-generated texts.

\begin{figure*}[!t]
    \centering
    \includegraphics[width=0.67\linewidth]{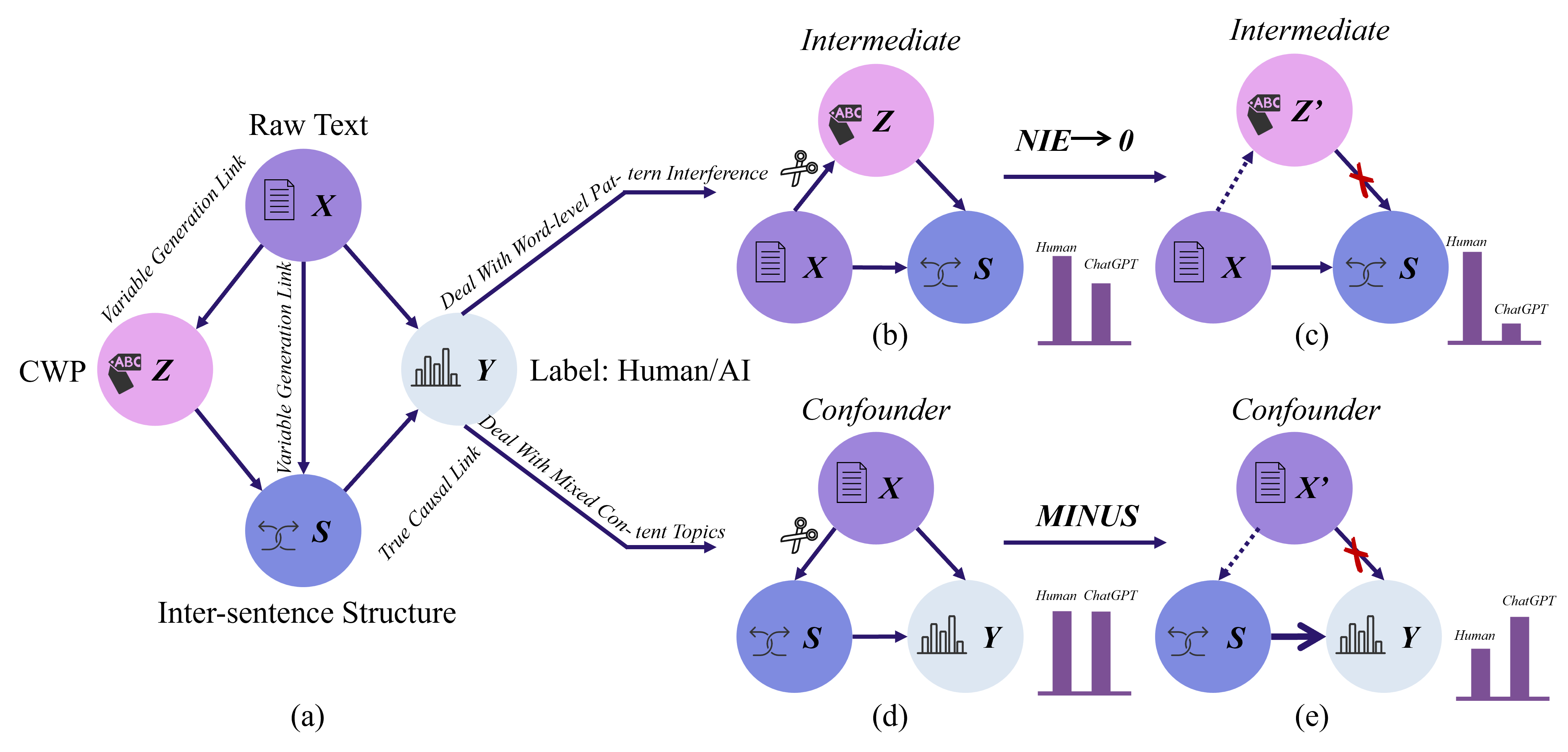}
    \caption{Causal graph of our text classification framework, illustrating the assumed relationships (a) and the counterfactual interventions used to decouple the inter-sentence structure $S$ from the word-level pattern $Z$ (b, c) and topic $X$ (d, e).}
    \label{fig:graph}
    \vspace{-5mm}
\end{figure*}

\subsection{Causal Reasoning and Inference}

Incorporating causal reasoning into NLP is an emerging and fascinating direction \cite{feder2022causal}, aiding in addressing  issues such as fairness (e.g., gender, race, education biases) \cite{lee2009advances,garg2019counterfactual}, interpretability, and data augmentation.

A prior study\cite{kaushik2019learning}conducted statistical analysis on the impact of gender and other emotion-independent factors on classification results in the IMDB dataset\cite{lin2011proceedings}, and addressed such biases by employing counterfactual data transformation. Another study \cite{veitch2021counterfactual} proposed a universally applicable causal graph for NLP domain classification, dividing text into three parts and establishing causal relationships among them. The authors also conducted theoretical analysis and introduced two regularization terms to guide detectors to be insensitive to spurious connections in different scenarios.

Moreover, in the field of computer vision, counterfactual learning has been applied to improve the model’s attention capture capability \cite{rao2021counterfactual} and reduce background biases in foreground classification tasks \cite{liu2022contextual, wang2021causal}. These applications provide valuable insights for addressing bias and improving robustness in AI-generated text detection tasks.

\section{Methods}
\subsection{Sentence Embedding Transformer Model}
As shown in \cref{fig:overview}, we aim to extract inter-sentence relationships and classify texts based on structural representations. Instead of Graph Convolutional Networks, which struggle to capture diverse and nuanced sentence relations, we simply employ Multi-Head Attention \cite{vaswani2017attention}, where each head models a distinct type of relationship (e.g., juxtaposition, continuity, transition) in a dedicated feature space.

We represent a text as a set of sentence embeddings:
\vspace{-5pt}
\begin{equation}
S = \mathcal{F}(X) = \{ S_{\text{cls}}, S_1, \ldots, S_M \}
\vspace{-5pt}
\end{equation}

where $\mathcal{F}$ denotes the pretrained RoBERTa-base, $S_i \in \mathbb{R}^{768}$, and $M = 32$. Each $S_i$ is obtained by averaging the token representations of all words in the $i^\text{th}$ sentence, thereby serving as its sentence-level embedding. A learnable position embedding is added to each sentence. We use the mean token embedding from the last layer of RoBERTa as the sentence representation, and include a special $\texttt{cls}$-token to aggregate global structural information. Texts shorter than $M$ sentences are padded and masked to avoid influencing attention.

For each attention head $h$, the sentence relation is computed as:
\vspace{-5pt}
\begin{equation}
\text{Relation}_{h} = \text{Softmax}\left( \frac{Q_h^T \cdot W_h}{\sqrt{d_{\text{model}}}} \right)
\vspace{-5pt}
\end{equation}
where $Q_h$ and $K_h$ are projections of the input embeddings. The structural representation from all heads is combined via:
\vspace{-5pt}
\begin{equation}
E_i = \text{Concat}_{h=1}^{N_\text{head}}(\text{Relation}_{h}\cdot V_h)
\vspace{-5pt}
\end{equation}

After processing through three transformer encoder layers, the updated representation set $E = \{E_{\text{cls}}, E_1, \dots, E_M\}$ is obtained. The final classification result is derived from the continuously aggregated $\texttt{cls}$-token:
\(
\text{Y}=\text{MLP}(E_\text{cls}).
\)

\subsection{Causal View of Classification Method}
To introduce causal reasoning into the text detection task, we adopt a Structural Causal Model (SCM) framework \cite{pearl2009causality}. Within this causal graph, we define $S$ as the variable representing inter-sentence structural information, $Z$ as the “ChatGPT-Style” word-level pattern (CWP), and $Y$ as the prediction label. As shown in \cref{fig:graph}(a), the causal structure includes the pathway $X \to Y$ representing the overall mapping from the input text to the label, and $S \to Y$ capturing the direct causal effect of inter-sentence relationships on the prediction. The link $Z \to Y$ reflects a spurious association where CWP directly influences the classification result via sentence embeddings, while $X \to (Z, S)$ indicates that both word-level patterns and inter-sentence structures are derived from the original text.

\subsection{Counterfactual Learning}

Conventional methods typically learn inter-sentence relationships through direct supervision of the final prediction $Y$, capturing structural information but lacking interpretability of how such structures causally affect the outcome. Counterfactual learning provides a solution to this issue. Based on the connection between $Z$ (word-level pattern) and $S$ (inter-sentence structure), we decompose the causal graph in \cref{fig:graph}(a) into two parts: \cref{fig:graph}(b) focuses on bias from CWP injection, and \cref{fig:graph}(d) addresses topic imbalance in limited training data.

Using causal intervention, denoted as $\text{do}(\cdot)$, we can analyze causal effects by fixing variables and cutting incoming links. For example, $\text{do}(Z = Z')$ sets $Z$ to a counterfactual value $Z'$ and removes the link $Z \rightarrow S$, blocking its influence on $S$.

To estimate the natural indirect effect (NIE) of $Z$ acting as a mediator in $X \rightarrow S$, we generate a counterfactual word-level style, $Z'$, by randomly replacing 30\% of verbs and nouns with their synonyms. This coarse paraphrasing is designed to disrupt the ChatGPT Word-level Pattern ($Z$), thereby weakening the model's reliance on superficial word choices. To ensure this intervention only alters the word-level style without changing the inter-sentence structure ($S$), the original position embeddings are preserved. This process is crucial for compelling the model to focus on the sentence-level structure, which we posit is the primary differentiator between human and AI-generated text. The NIE is then computed as:

\vspace{-5pt}
\begin{equation}
Y_{\text{NIE}} = \mathbb{E}_{Z' \sim \gamma} \left[ Y(\text{do}(Z = Z'), X = \mathbf{X}) - Y(Z = \mathbf{Z}, X = \mathbf{X}) \right]
\vspace{-5pt}
\end{equation}
and incorporated via a regularization term:
\vspace{-5pt}
\begin{equation}
\mathcal{L}_{\text{NIE}} = \mathcal{L}_{\text{BCE}}(Y_{\text{NIE}}, Y)
\vspace{-5pt}
\end{equation}
Similarly, for the direct effect (DE) of $S \rightarrow Y$, we aim to decouple the structural effect from topic-related bias. This is achieved by generating new content $X'$ on a different topic, while holding the inter-sentence structure S constant:
\vspace{-5pt}
\begin{equation}
Y_{\text{DE}} = \mathbb{E}_{X' \sim \gamma'} \left[ Y(X = \mathbf{X}, S = \mathbf{S}) - Y(\text{do}(X = \mathbf{X}'), S = \mathbf{S}) \right]
\vspace{-5pt}
\end{equation}
with a corresponding loss:
\vspace{-5pt}
\begin{equation}
\mathcal{L}_{\text{DE}} = \mathcal{L}_{\text{BCE}}(Y_{\text{DE}}, Y)
\vspace{-5pt}
\end{equation}

The total loss combines the standard classification loss with counterfactual regularization terms:
\vspace{-5pt}
\begin{equation}
\mathcal{L} = \mathcal{L}_{\text{BCE}}(Y, \text{Label}) + \mathcal{L}_{\text{NIE}} + \mathcal{L}_{\text{DE}}
\vspace{-5pt}
\end{equation}
This approach enhances the model's focus on structural rather than lexical or topical features, improving robustness to word-level variations and content shifts.

\section{Experiment}
\label{sec:typestyle}

\subsection{Baseline and Experiment Setup}

To establish robust baselines, we follow the methodology of prior work\cite{ guo2023close}. Our primary baseline, RoBERTa-HC3, fine-tunes the respective roberta-base\cite{liu2019roberta} (for English) and hfl/chinese-roberta-wwm-ext\cite{cui2020revisiting} (for Chinese) models on the public HC3 dataset for 2 epochs. We also introduce an enhanced baseline, RoBERTa-HC3FT, which undergoes further training on an augmented HC3 set with synonym substitutions to test the limits of word-level adaptation.

Our proposed model is designed upon RoBERTa architectures. After pre-training on HC3, it is fine-tuned for 2 epochs on our curated dataset, which consists of 9,506 English scientific abstracts and a collection of life FAQs (175,524 English; 53,415 Chinese) structured in semantically similar groups of 3-4. This stage integrates causal inference and counterfactual learning to enhance its understanding of intrinsic textual structures. All models were trained with a $5 \times 10^{-5}$ learning rate and a batch size of 16. Our evaluation for each task utilizes 25,049 English and 7,696 Chinese test samples. The evaluation covers the following tasks: HC3 (standard detection benchmark), Cyclical Translation (semantic-invariant robustness test), Substitution(lexical substitutions mimicking user polishing), and Any Alteration (combined modifications for overall robustness).

\subsection{Results and Analysis}
\subsubsection{Main Performance Evaluation}
We evaluate all models on a series of tasks designed to test their core detection capabilities and robustness against common alterations. The primary results for all three models on the English and Chinese test sets are presented in Table~\ref{tab:eng_main} and Table~\ref{tab:chn_main}, respectively, with Accuracy as the metric. 
\begin{table}[H]
\centering
\vspace{-3mm}
\caption{Model Performance on English Main Test Sets (Accuracy \%)}
\label{tab:eng_main}
\scriptsize
\begin{tabular}{lcccc}
\toprule
\textbf{Model} & \textbf{HC3} & \textbf{Translation} & \textbf{Substitution} & \textbf{Any Alteration} \\
\midrule
RoBERTa-HC3\cite{ guo2023close}   &\textbf{99.88} & 87.15 & 95.74 & 88.64 \\
RoBERTa-HC3FT & 99.59 & 94.40 & 99.03 & 94.05 \\
\textbf{Ours}      & 99.60 & \textbf{98.57} & \textbf{99.43} & \textbf{98.36} \\
\bottomrule
\end{tabular}
\vspace{-5 mm}
\end{table}

\begin{table}[H]
\centering
\vspace{-3mm}
\caption{Model Performance on Chinese Main Test Sets (Accuracy \%)}
\label{tab:chn_main}
\scriptsize
\begin{tabular}{lcccc}
\toprule
\textbf{Model} & \textbf{HC3} & \textbf{Translation} & \textbf{Substitution} & \textbf{Any Alteration} \\
\midrule
CN-RoBERTa-HC3\cite{ guo2023close}   & 96.91 & 80.18 & 95.47 & 85.90 \\
CN-RoBERTa-HC3FT & 99.21 & 89.27 & 98.73 & 93.27 \\
\textbf{Ours}   & \textbf{99.49} & \textbf{94.80} & \textbf{99.27} & \textbf{96.32} \\
\bottomrule
\end{tabular}
\vspace{-5 mm}
\end{table}

As shown in Table~\ref{tab:eng_main} and Table~\ref{tab:chn_main}, the baseline RoBERTa-HC3 model performs well on the original HC3 task but declines significantly under semantic alterations, confirming the brittleness of standard detectors. The enhanced RoBERTa-HC3FT model improves robustness through exposure to lexical diversity, yet remains limited on complex structural tasks like Translation. In contrast, our proposed model consistently achieves the highest accuracy across all tasks, with a notable advantage on challenging subsets such as ``Any Alteration" and ``Translation", demonstrating its superior generalization and structural reasoning capability.

\subsubsection{Domain Generalization Evaluation}

To assess the model's generalization capabilities, we evaluated their performance on the ``Any Alteration" task across multiple vertical domains. We use the F1-score as the metric for this evaluation. The results are presented in Table~\ref{tab:eng_domain} for English and Table~\ref{tab:chn_domain} for Chinese.

\begin{table}[H]
\centering
\vspace{-3 mm}
\caption{Domain-Specific Generalization on English Datasets (F1-Score)}
\label{tab:eng_domain}
\scriptsize
\begin{tabular}{lccc}
\toprule
\textbf{Domain}          & \textbf{RoBERTa-HC3} & \textbf{RoBERTa-HC3FT} & \textbf{Ours} \\
\midrule
Finance                  & 0.9265                 & 0.9405                   & \textbf{0.9788}   \\
Medicine                 & 0.9555                 & 0.9214                   & \textbf{0.9913}   \\
Reddit(ELI5)             & 0.8835                 & 0.9429                   & \textbf{0.9882}   \\
Wikipedia (CS/AI)        & 0.8226                 & 0.8366                   & \textbf{0.9256}   \\
\bottomrule
\end{tabular}
 \vspace{-5mm}
\end{table}

\begin{table}[H]
\centering
\vspace{-3 mm}
\caption{Domain-Specific Generalization on Chinese Datasets (F1-Score)}
\label{tab:chn_domain}
\scriptsize
\begin{tabular}{lccc}
\toprule
\textbf{Domain}          & \textbf{CN-RoBERTa-HC3} & \textbf{CN-RoBERTa-HC3FT} & \textbf{Ours} \\
\midrule
Finance                  & 0.8999                 & 0.9441                   & \textbf{0.9829}    \\
Medicine                 & 0.7344                 & 0.8726                   & \textbf{0.9486}    \\
Law                      & 0.8623                 & 0.9609                   & \textbf{0.9929}    \\
Baike                    & 0.7399                 & 0.8116                   & \textbf{0.8873}    \\
psychology               & 0.8451                 & 0.9360                   & \textbf{0.9687}    \\
\bottomrule
\end{tabular}
\vspace{-4.5 mm}
\end{table}

The results in Table~\ref{tab:eng_domain} and Table~\ref{tab:chn_domain} clearly indicate that our model possesses superior generalization ability. This suggests that by learning the fundamental structural properties of text rather than surface-level statistics, our model is more adaptable and reliable when faced with content from unseen domains.

\section{CONCLUSION}
Our work demonstrates and analyzes the pervasive critical vulnerability of conventional AIGC detectors. Our experiments confirm that they perform well on original AI-generated text but significantly fail when faced with common semantic modifications like synonym substitution and cyclical translation. To address this, we propose a novel lightweight, sentence-level detector that leverages causal inference to analyze invariant deep textual structures, achieving state-of-the-art robustness and generalization. This highlights that structural analysis is a more reliable and effective direction for AIGC detection. We will soon release the large-scale language benchmark datasets used in this study to support the community and foster future researches.

\bibliographystyle{IEEEtran}

\bibliography{reference}

\end{document}